\documentclass[10pt,twocolumn,letterpaper]{article}

\usepackage[final]{cvpr}      
\pdfoutput=1
\usepackage[mathscr]{eucal}
\usepackage[table]{xcolor}

\usepackage[pagebackref,breaklinks,colorlinks]{hyperref}
\usepackage[accsupp]{axessibility}
\usepackage{CJKutf8}
\usepackage{graphicx}
\usepackage{float}
\usepackage{array}
\usepackage{amsmath,amssymb,amsfonts}
\usepackage{booktabs}
\usepackage{tabularx}
\usepackage{multirow}
\usepackage{wrapfig}
\usepackage{paralist} 

\usepackage[accsupp]{axessibility}

\usepackage[capitalize]{cleveref}
\crefname{section}{Section}{Secs.}
\Crefname{section}{Section}{Sections}
\Crefname{table}{Table}{Tables}
\crefname{table}{Table}{Tabs.}

\def\cvprPaperID{5377} 
\def\confName{CVPR}
\def\confYear{2023}

\begin{document}

\title{MotionTrack: Learning Robust Short-term and Long-term Motions\\ for Multi-Object Tracking}

\author{Zheng Qin$^{1 \dag}$ ~Sanping Zhou$^{1 \dag}$ ~Le Wang$^{1 *}$ ~Jinghai Duan$^2$ ~Gang Hua$^3$~Wei Tang$^4$ \\
	$^{1}$National Key Laboratory of Human-Machine Hybrid Augmented Intelligence, \\ National Engineering Research Center for Visual Information and Applications,  \\Institute of Artificial Intelligence and Robotics, Xi'an Jiaotong University\\
	$^{2}$School of Software Engineering, Xi'an Jiaotong University\\
	$^{3}$Wormpex AI Research ~
	$^{4}$University of Illinois at Chicago\\
}

\maketitle
\footnotetext{$^\dag$Co-first authors. $^*$Corresponding author.}
\begin{abstract}
The main challenge of Multi-Object Tracking~(MOT) lies in maintaining a continuous trajectory for each target. Existing methods often learn reliable motion patterns to match the same target between adjacent frames and discriminative appearance features to re-identify the lost targets after a long period. However, the reliability of motion prediction and the discriminability of appearances can be easily hurt by dense crowds and extreme occlusions in the tracking process. In this paper, we propose a simple yet effective multi-object tracker, i.e., MotionTrack, which learns robust short-term and long-term motions in a unified framework to associate trajectories from a short to long range. For dense crowds, we design a novel Interaction Module to learn interaction-aware motions from short-term trajectories, which can estimate the complex movement of each target. For extreme occlusions, we build a novel Refind Module to learn reliable long-term motions from the target's history trajectory, which can link the interrupted trajectory with its corresponding detection. Our Interaction Module and Refind Module are embedded in the well-known tracking-by-detection paradigm, which can work in tandem to maintain superior performance. Extensive experimental results on MOT17 and MOT20 datasets demonstrate the superiority of our approach in challenging scenarios, and it achieves state-of-the-art performances at various MOT metrics. 
\end{abstract}

\section{Introduction}
\label{sec:intro}

Multi-Object Tracking~(MOT) is a fundamental task in computer vision, which has a wide range of applications, such as autonomous driving~\cite{driving} and intelligent surveillance~\cite{surveil}. It aims at jointly locating targets through bounding boxes and recognizing their identities throughout a
whole video~\cite{regression1}. Though great progress has been made in the past few years, MOT still remains a challenging task due to the dynamic environment, such as dense crowds and extreme occlusions, in the tracking scenario. 

\begin{figure}[t]
  \centering
  \includegraphics[width=\linewidth]{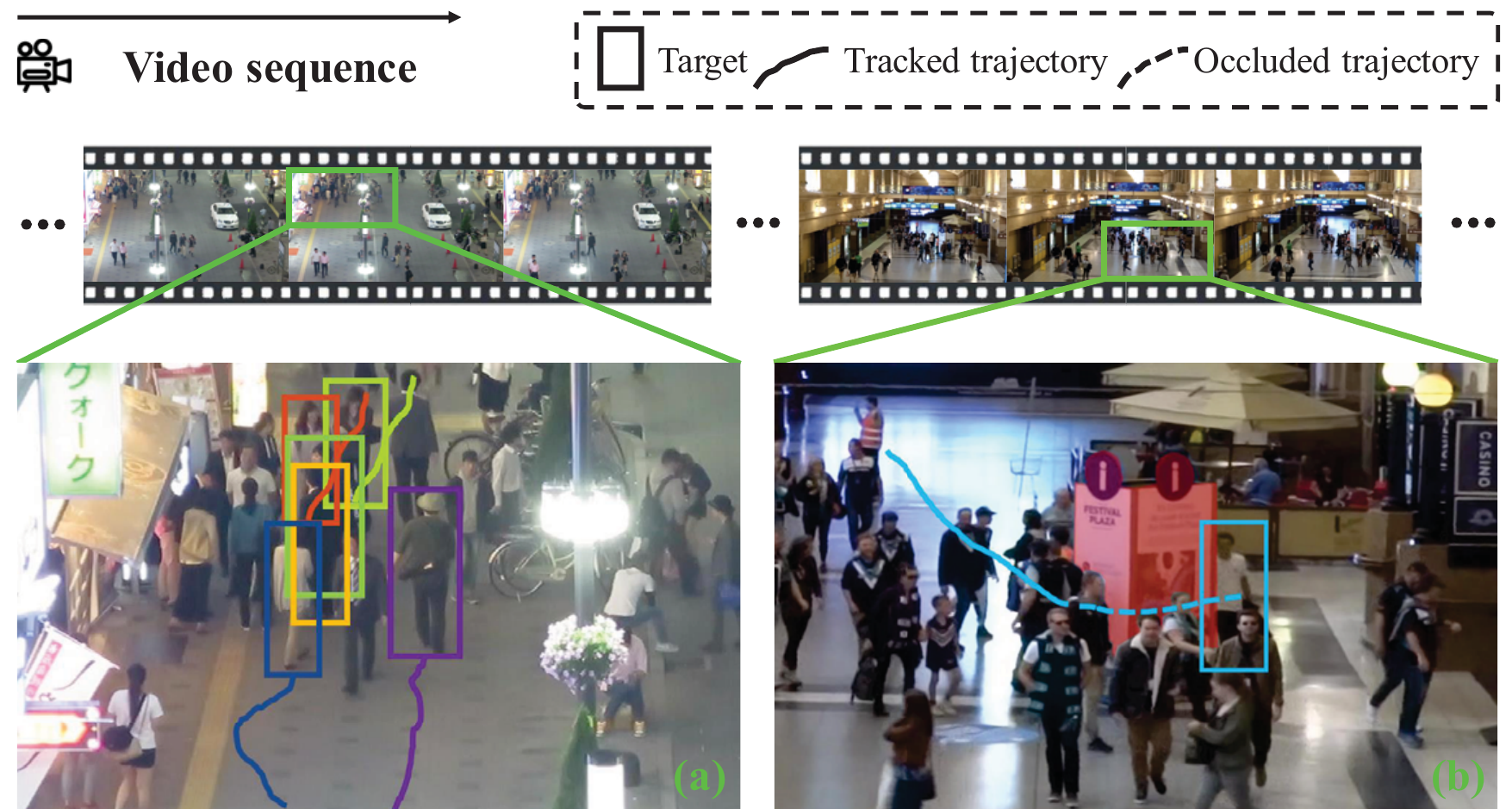}
   \caption{\textbf{Illustration of challenging scenarios in different videos}. (a)~\textbf{Dense crowds.} Pedestrians do not move independently in this situation. They will be affected by their surrounding neighbors to avoid collisions which will make their motion patterns hard to learn in practice. (b)~\textbf{Extreme occlusion.} Pedestrians are easily occluded by fixed facilities for a long period, such as billboard and sunshade, in which the dynamic environment will make them undergone a large appearance variation.}
   \label{fig:intro}
\end{figure}

In general, the existing MOT methods either follow the tracking-by-detection~\cite{tracking-by-detection} or tracking-by-regression~\cite{centertrack,regression1,regression2}, paradigm. The former methods first detect objects in each video frame and then associate detections between adjacent frames to create individual object tracks over time. The latter methods conduct tracking differently: the object detector not only provides frame-wise detections but also replaces the data association with a continuous regression of each tracklet to 
its new position. Regardless of the paradigm, all methods need to address the short-range and long-range association problems, \emph{i.e.}, how to associate the alive tracklets with detections in a short time, and how to re-identify the lost tracklets with detections after a long period.


For the short-range association problem, discriminative motion patterns and appearance features~\cite{sort,deepsort} are often learned to conduct data association between adjacent frames. However, as shown in Figure~\ref{fig:intro}~(a), it is tough to learn discriminative representations in the dense crowd scenario. On the one hand, the bounding boxes of detections are too small to be distinguished by their appearances. On the other hand, different targets need to plan suitable paths to avoid collisions, which makes the resulting motions very complex in the tracking process. 
For the long-range association problem, prior works~\cite{deepsort,wang2020towards,cstrack} usually learn discriminative appearance features to re-identify the lost targets after long occlusion~\cite{zhoureid1,zhoureid2,zhoureid3}. As shown in Figure~\ref{fig:intro}~(b), the main bottleneck of these methods is how to keep the robustness of features against different poses, low resolution, and poor illumination for the same target. To alleviate this issue, the memory technology~\cite{memot, Towards-Discriminative} is widely applied to store diverse features for each target to match different targets in a multi-query manner. Moreover, a lot of memories and time will be consumed by the memory module and multi-query regime, unfriendly to real-time tracking. 

In this paper, we propose a simple yet effective object tracker, \emph{i.e.}, MotionTrack, to address the short-range and long-range association problems in MOT. In particular, our MotionTrack follows the tracking-by-detection paradigm, in which both interaction-aware and history trajectory-based motions are learned to associate trajectories from a short to long range. To deal with the short-range association problem, we design a novel Interaction Module to model all interactions between targets, which can predict their complex motions to avoid collisions. The Interaction Module uses an asymmetric adjacency matrix to represent the interaction between targets, and obtains the prediction after the information fusion by a graph convolution network. Thanks to the captured target interaction, those short-term occluded targets can be successfully tracked in dense crowds. To deal with the long-range association problem, we design a novel Refind Module based on the history trajectory of each target. It can effectively re-identify the lost targets
through two steps: correlation calculation and error compensation. For the lost tracklets and the unmatched detections, the correlation calculation step takes the features of history trajectories and current detections as input, and computes a correlation matrix to represent the possibility that they are associated. Afterward, the error compensation step is further taken to revise the occluded trajectories. Extensive experiments on two benchmark datasets (MOT17 and MOT20) demonstrate that our proposed MotionTrack outperforms the previous state-of-the-art methods.
\begin{figure*}[t]
\centering
   \includegraphics[width=1\linewidth]{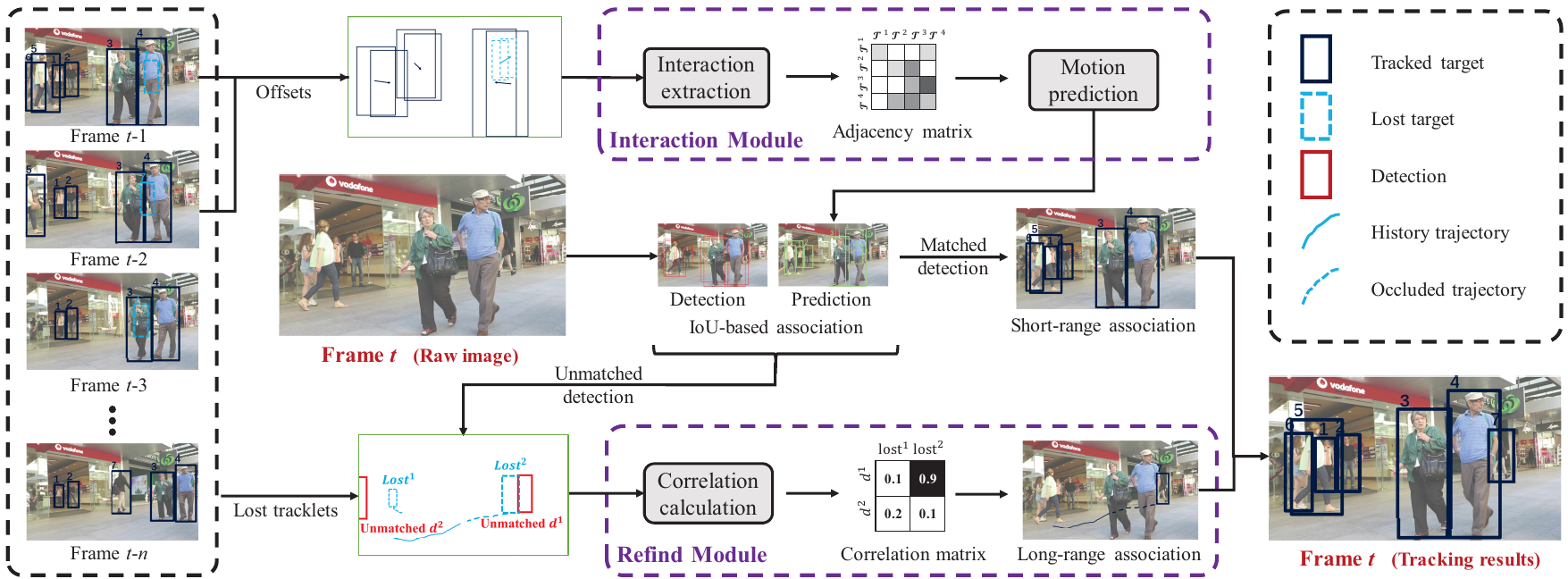}
   \caption{{\bf Overview of our MotionTrack}. The {\bf Interaction Module} captures the directional interaction relationship between tracklets, then fuses the interaction information and predicts the location in the next frame for the short-range association. The {\bf Refind Module} analyzes the correspondence between unmatched lost tracklets and detections by correlation calculation, then the matched pairs are further chosen to complete the long-range association via an additional error compensation. Finally, the short-range and long-range associations are combined to generate complete tracking results.}
   \label{fig:Overview}
\end{figure*}


The main contribution of this work can be highlighted as follows:~(1) We propose a simple yet effective multi-object tracker, MotionTrack, to address the short-range and long-range association problems;~(2) We design a novel Interaction Module to model the interaction between targets, which can handle complex motions in dense crowds;~(3)~We design a novel Refind Module to learn discriminative motion patterns, which can re-identify the lost tracklets with current detections.


\section{Related Work}
\label{sec:RelatedWork}
\noindent {\bf Tracking-by-Detection.}
The recent advancement of object detection brings remarkable improvement to the tracking-by-detection paradigm~\cite{tracking-by-detection}. 
In this framework, an existing object detector~\cite{faster-rcnn,ge2021yolox} generates detections in the current frame, then a matching algorithm, \emph{e.g.}, the Hungarian algorithm~\cite{munkres1957algorithms}, builds tracklets by associating the detections across different frames. Many efforts have been made in different aspects to improve the effectiveness of this paradigm. For example,
SORT~\cite{sort} adds the Kalman Filter~(KF)~\cite{kalman1960new} to approximate
the inter-frame displacements, while other models~\cite{deepsort,wang2021multiple,pang2021quasi,yu2022towards} focus on distinguishing appearance features to improve the matching accuracy. Some other works~\cite{braso2020learning,kim2015multiple,he2021learnable} formulate data association as a graph optimization problem by considering each detection as a graph node. For example, ByteTrack~\cite{bytetrack} enhances tracking performance by fully using detections. We follow this tracking-by-detection paradigm and propose a novel framework to extract interaction-aware and  history trajectory-based motions for more accurate short and long range association.

\noindent{\bf Motion Models.} The motion models can be divided into filter-based and model-based methods in MOT. The filter-based methods mainly consider motion prediction as state estimation. For example, the well-known SORT~\cite{sort} introduces KF~\cite{kalman1960new} into MOT as a linear constant velocity model based on the assumption of independence across the objects and the camera motion, which inspires a series of works ~\cite{fairmot,Motion-aware,Detecting-invisible-people} to improve the motion prediction in different aspects. Later, many works~\cite{tracktor,strongsort,botsort,modelling-ambiguous} consider the camera motion compensation of the tracker for robust tracking. Meanwhile, some works~\cite{strongsort,giaotracker} using KF variants to further improve prediction accuracy. 

Recently, model-based methods combine motion and visual information to provide better predictions based on a data-driven paradigm. For example, Tracktor~\cite{tracktor} adopts the regression part from Faster R-CNN~\cite{faster-rcnn} to predict the displacement of targets between adjacent frames. FFT~\cite{fft} further adds optical flow to help regress the displacement. Besides, CenterTrack~\cite{centertrack} builds a tracking branch to predict the motion specifically. ArTIST~\cite{probabilistic} considers motion as a probability distribution and implicitly models all surrounding interactions with max-pooling 
to one feature. However, most motion models do not consider explicit interactions between targets, especially in dense crowd scenes. As a result, they cannot accurately estimate the complex movements of nearby targets.

\noindent{\bf Occlusion.} How to deal with occlusions has been a long-standing challenge in MOT. In particular, occlusions can be divided into short-term and long-term occlusions. The short-term occlusion means the target is incomplete in a frame as some other objects occlude it. It prevents the extraction of high-quality detection features for the association. To address this issue, some works try to separate the unoccluded and occluded targets~\cite{Towards-Discriminative,Online-multiple}. 

The long-term occlusion occurs when the tracking target is lost for a long period due to obstacles. To alleviate this issue, DeepSORT~\cite{deepsort} proposes a cascaded matching strategy that first matches the detection boxes to the alive tracklets and then to the lost targets based on appearance features. MeMOT~\cite{memot} builds a memory bank to store the appearance features of the tracklets for retrieval. QuoVadis~\cite{dendorfer2022quo} convert trajectories~\cite{socialgan,shi2021sgcn} to a bird's eye view. ByteTrack~\cite{bytetrack} re-identifies the lost tracklet using the IoU score between the tracklet's iterative prediction and detection. 
However, these methods will become less reliable as the occlusion time becomes longer. We propose a new representation of the tracklet based on its history trajectory, so that the model can still provide a reliable matching strategy in extremely occluded scenes.

\section{Method}
\label{sec:Method}
\subsection{Notation}
As shown in Figure~\ref{fig:Overview}, our MotionTrack  follows a well-known tracking-by-detection paradigm~\cite{bytetrack}. We first process each frame with YOLOX~\cite{ge2021yolox} to obtain the detection results. The detections are denoted as $\mathcal{D}^{t} = \{\mathbf{d}_{i}^{t}\}_{i=1}^N$ containing $N$ detections in frame~$t$. A detection $\mathbf{d}^{t}_{i}\in\mathbb{R}^4$ is represented as $(x, y, w, h)$, where $(x, y)$ means the bounding box center, and $w$ and $h$ indicate its width and height, respectively. We denote the set of $M$ tracklets by $\mathbb{T} = \{\mathcal{T}_{j}\}_{j=1}^M$. $\mathcal{T}_{j}$ is a tracklet with identity $j$ and is defined as $\mathcal{T}_{j}$ = \{$\mathbf{l}^{t_0}_{j}$, $\mathbf{l}^{t_0+1}_{j}$, ..., $\mathbf{l}^{t}_{j}$\}, where $\mathbf{l}^{t}_{j}$ is the location in frame~$t$, and $t_0$ is the initialized moment.

When tracking begins, we initialize the set of tracklets $\mathbb{T}$ with  $\mathcal{D}^{1}$. 
For the subsequent video frames, we assign the new detections to their corresponding tracklets, and update $\mathbb{T}$ at each time step. Throughout the whole video sequence, new tracklets are constantly initialized and incorporated into $\mathbb{T}$. Meanwhile, the existing tracklets may be terminated and removed from $\mathbb{T}$. In the tracking process, the tracklet may be interrupted, for example due to dense crowds and extreme occlusion, therefore we further divide $\mathbb{T}$ into two parts,~\emph{i.e.}, $\mathbb{T}^\text{alive}$ and $\mathbb{T}^\text{lost}$, which denote the set of tracked tracklets and the set of lost but not yet removed tracklets, respectively. 
A tracklet in $\mathbb{T}^\text{alive}$ is represented as $\mathcal{T}_{k}$ = \{$\mathbf{d}^{t_0}_{k}$, $\mathbf{d}^{t_0+1}_{k}$, ...,$\mathbf{d}^{t}_{k}$\}, and a tracklet in $\mathbb{T}^\text{lost}$ is represented as $\mathcal{T}_{s}$ = \{$\mathbf{d}^{t_0}_{s}$, $\mathbf{d}^{t_0+1}_{s}$, ...,$\mathbf{p}^{t_{\text{lost}}}_{s}$,$\mathbf{p}^{t_{\text{lost}}+1}_{s}$, ...,$\mathbf{p}^{t}_{s}$\}, where $t_{\text{lost}}$ denotes the moment it is lost,  and $\mathbf{p}^{t}_{s}$ is the prediction during occlusion calculated in~\cref{IModule} below.

\subsection{Overview of MotionTrack}
Our MotionTrack mainly executes two steps for each video frame: 
\begin{itemize}
	\item {\bf Step 1: Short-range association.} Modeling the inter-tracklet interaction to obtain more accurate predictions and the short-range tracking results.
	\item {\bf Step 2: Long-range association.} Re-identifying lost tracklets based on the history trajectory and unmatched detections and then compensating the trajectory during occlusion.
\end{itemize}

\begin{figure}[t]
  \centering
  \includegraphics[width=0.9\linewidth]{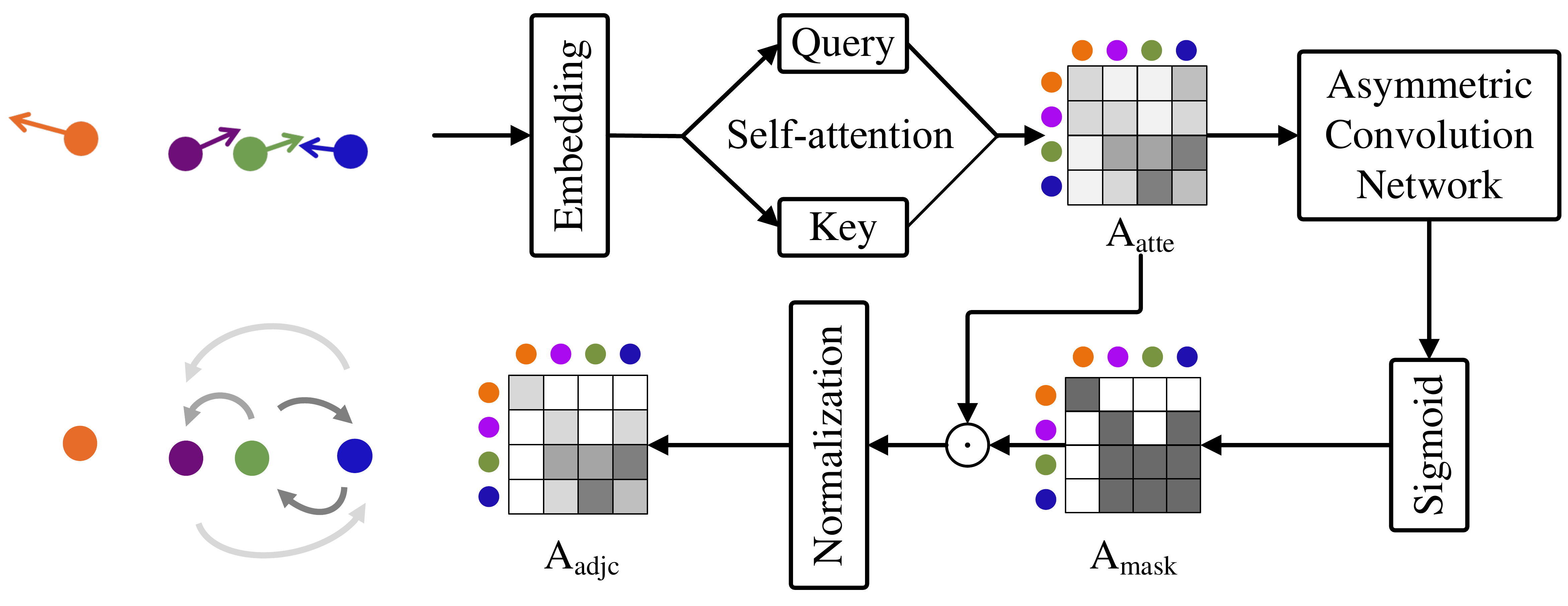}
   \caption{{\bf Illustration of Interaction Extraction}. We capture interactions between tracklets, and the shades of arrows reflect the degree of interaction.}
   \label{fig:Interaction-module}
\end{figure}

All tracks go through three states from birth to death: alive, lost, and dead. The tracklet being tracked is called alive, and its state is converted to lost upon interruption by accident, such as extreme occlusion. 
When the interruption occurs for a long time, the tracklet becomes dead, and we remove it from the tracklets set $\mathbb{T}$.  

\noindent {\bf Short-range Association.}
Given the input video frame~$t$, we first obtain its detections  $\mathcal{D}^{t}$. In addition, we have obtained the set of $M$ tracklets $\mathbb{T}$ up to frame~$t-1$, with $S$ lost tracklets $\mathbb{T}^{\text{lost}}$. 
Then, we construct the directed interactions between tracklets to get the predictions at frame~$t$. 
As shown in Figure~\ref{fig:Overview}, we first calculate the iterative offsets $\mathbf{O}^{t}\in\mathbb{R}^{M\times4}$, where each row $\mathbf{o}^{t}_{j}=(\Delta x{_j^t}, \Delta y{_j^t}, \Delta w{_j^t}, \Delta h{_j^t})$ and $\Delta$ denotes the offset from frame~$t-2$ to frame~$t-1$. The Interaction Module concatenates absolute coordinates and offsets as input, denote as $\mathbf{I}^{t}~\in~\mathbb{R}^{M\times8}$.
An asymmetric interaction matrix $\mathbf{A}^{\text{adjc}}  \in  [0,1]^{M \times M}$ is obtained through interaction extraction, where each element indicates the impact of one tracklet on the other. Afterward, $\mathbf{A}^{\text{adjc}}$ is used to estimate an accurate offset in the motion prediction step. Finally, we follow the association policy in ByteTrack~\cite{bytetrack} to update the alive tracklets $\mathbb{T}^\text{alive}$ with matched detections and record the predictions for lost tracklets $\mathbb{T}^\text{lost}$.

\noindent {\bf Long-range Association.}
Suppose that there are $S$ lost tracklets and $U$ unmatched detections after the short-range association. Some unmatched detections could share the identities with those lost tracklets, which motivates us to learn history trajectory-based representations for the long-range association. In particular, we first calculate the correlation between the $S$ lost tracklets and the $U$ unmatched detections based on the spatial distribution of trajectories and the velocity-time relationship to obtain the correlation matrix $\mathbf{C}^{\text{corre}}  \in  [0,1]^{S \times U}$. Then, we retain highly relevant pairs and utilize error compensation to refine the trajectory. Finally, we combine the results of short-range and long-range associations to generate the complete tracking results at frame~$t$.

\subsection{Interaction Module}
\label{IModule}
To obtain more accurate tracklets, we capture the directed interactions between tracklets in the interaction extraction step and then use them to estimate the offsets between two consecutive frames in the motion prediction step.

\noindent {\bf Interaction Extraction.}
As shown in Figure~\ref{fig:Interaction-module}, we first obtain the attention matrix $\mathbf{A}^\text{atte} \in \mathbb{R}^{M \times M}$, which measures the interaction magnitude between every pair of tracklets, using the self-attention mechanism~\cite{vaswani2017attention}:
\begin{equation}
\begin{aligned}
\mathbf{E}^{t} &= \phi\left( \mathbf{I}^{t}, \mathbf{W}^\text{E} \right), \\
\mathbf{Q}^{t} &= \phi\left(\mathbf{E}^{t}, \mathbf{W}^\text{Q} \right), \\
\mathbf{K}^{t} &= \phi\left(\mathbf{E}^{t}, \mathbf{W}^\text{K}\right), \\
\mathbf{A}^{\text{atte}} &=\operatorname{Softmax}\left(\frac{\mathbf{Q}^{t} {\mathbf{K}^{t}}^{\text{T}}}{\sqrt{d_{\mathrm{ }}}}\right),
\end{aligned}
\end{equation}
where $\phi (\cdot, \cdot)$  denotes linear transformation (multiplying a weight matrix and adding a bias),  $\mathbf{E}^{t}$ is a higher-dimension embedding mapped from $\mathbf{I}^{t}$. $\mathbf{Q}^{t} \in \mathbb{R}^{M \times D}$ and $\mathbf{K}^{t} \in \mathbb{R}^{M \times D}$ are the query and key of the self-attention mechanism. $\mathbf{W}^\text{E}$, $\mathbf{W}^\text{Q}$, and $\mathbf{W}^\text{K}$ are the weights of the linear transformation, and $\sqrt{d}  = \sqrt{D}$ is the scaling factor~\cite{vaswani2017attention}.

We use the attention matrix $\mathbf{A}^\text{atte}$ to express the asymmetric interaction between different tracklets instead of the undirected spatial distance.
The $(i, j)$-th element in $\mathbf{A}^\text{atte}$ represents the influence of tracklet $i$ on tracklet $j$. 
To further consider the whole scene, such as group behavior, we model higher levels of interaction via a cascade of asymmetric convolution~\cite{ding2019acnet}:
\begin{equation}
\begin{aligned}
&\mathbf{A}_{l} = \delta\left(\rm{conv}\left(\mathbf{A}_\textit{l-1}, \mathbf{K}_{\mathrm{1 \times \kappa}}\right) + \rm{conv}\left(\mathbf{A}_\textit{l-1}, \mathbf{K}_{\mathrm{\kappa \times 1}}\right)\right),  \\
\end{aligned}
\end{equation}
where $\mathbf{K}_{1 \times \kappa}$ and $\mathbf{K}_{\kappa \times 1}$ are the asymmetric convolution kernels, $\delta$ denotes the PReLU, and $\mathbf{A}_{\mathrm{0}}$ is initialized as $\mathbf{A}^{\text{atte}}$.
There are $L$ convolution layers.
Afterward, to capture significant interactions between tracklets, we retain only high attention values in $\mathbf{A}^{\text{adjc}} \in [0,1]^{M \times M}$:
\begin{equation}
\begin{aligned}
\mathbf{A}^{\text{mask}} &= \rm{sgn}\left(\varphi(\mathbf{A}_{\mathrm{L}}) - \xi \right),\\
\mathbf{A}^{\text{adjc}}\ &= \mathbf{A}^{\text{mask}} \odot \mathbf{A}^{\text{atte}},
\end{aligned}
\end{equation}
where $\varphi$ and $\odot$ denote the sigmoid function and element-wise multiplication, respectively,
$\rm{sgn}$ is a sign function, and $\xi \in[0, 1]$ is a threshold.
Finally, we normalize all the non-zero elements in $\mathbf{A}^{\text{adjc}}$. 

\noindent {\bf Motion Prediction.}
After extracting the interaction matrix between tracklets, we use a graph convolution~\cite{kipf2016semi} to fuse the interactions for each tracklet and get the prediction using a multi-layer perceptron~($\rm{MLP}$):
\begin{equation}
\begin{aligned}
&\mathbf{P}^{\text{offs}} =\rm{MLP} \left(\delta \left(\phi(\mathbf{A}^{\text{adjc}}\cdot\mathbf{O}^\textit{t},\mathbf{W}^\text{G})\right)\right),\\
\end{aligned}
\end{equation}
where $\mathbf{W}^\text{G}$ is the weight of the linear transformation. The prediction  $\mathbf{P}^{\text{offs}} \in \mathbb{R}^{M \times 4}$  will be used with $\mathcal{D}^{t}$ for IoU-based association.

\begin{figure}[t]
  \centering
  \includegraphics[width=0.9\linewidth]{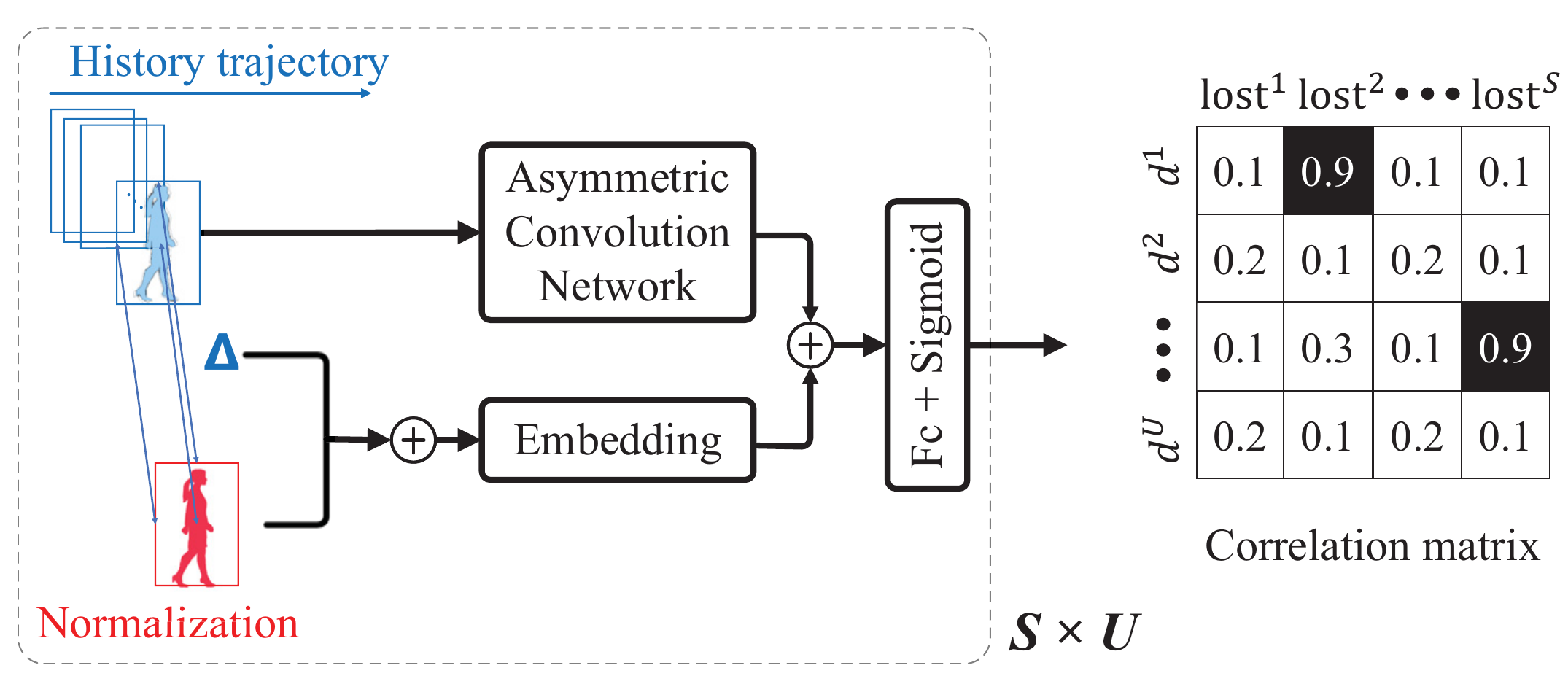}
   \caption{{\bf Illustration of Correlation Calculation}. We compute a correlation matrix to represent the association probabilities between lost tracklets and detections.}
   \label{fig:Refind Module}
\end{figure}

\subsection{Refind Module}
To refind the lost tracklet, we first identify its matched detection in the correlation calculation step and then refine the occluded trajectory in the error compensation step.

\noindent {\bf Correlation Calculation.}
After IoU-based association, there are $U$ unmatched detections $\mathbf{D}^{\text {rest}} \in \mathbb{R}^{U \times 5}$, and $S$ lost tracklets $\mathbf{T}^{\text {lost}} \in \mathbb{R}^{S \times 30 \times 5}$, in which we record the last thirty alive locations for each lost tracklet.
$\mathbf{D}^{\text {rest}}$ and $\mathbf{T}^{\text {lost}}$ include both time and locations, \emph{i.e.},  $(t, x, y, w, h)$, and they are the inputs to the Refind Module.
As shown in Figure~\ref{fig:Refind Module}, we first normalize $\mathbf{D}^{\text {rest}}$ and $\mathbf{T}^{\text {lost}}$ in the last dimension and then extract features from them separately.
We apply the asymmetric convolution to $\mathbf{T}^{\text {lost}}$ in the second and third dimensions, respectively, and then pool them into feature vectors $\mathbf{F}^{\text {traj}} \in \mathbb{R}^{S \times D}$:
\begin{equation}
\begin{aligned}
\mathbf{T}_{l}& = \delta\left(\rm{conv}(\mathbf{T}_\textit{l-1}, \mathbf{K}_{\mathrm{\kappa \times 1}})\right),   \\
\mathbf{F}^\text{traj} &= \rm{pool}\left(\delta ( \rm{conv}(\mathbf{T}_{L}, \mathbf{K}_{\mathrm{1 \times \kappa}}))\right),   \\
\end{aligned}
\end{equation}
where $\mathbf{T}_{\mathrm{0}}$ is initialized as $\mathbf{T}^{\text{lost}}$ and there are $L$ convolution layers.
We calculate the difference between the detection and the last alive location, and concatenate it with the detection as $\hat{\mathbf{D}^{\text {rest}}} \in \mathbb{R}^{(S \times U) \times 10}$. Then, we map it to high-dimensional features $\mathbf{F}^{\text {dete}} \in \mathbb{R}^{(S \times U) \times D}$ as follows:
\begin{equation}
\begin{aligned}
&\mathbf{F}^\text{dete} =\phi(\hat{\mathbf{D}^{\text {rest}}},\mathbf{W}^\text{D}),
\end{aligned}
\end{equation}
where $\mathbf{W}^\text{D}$ is the weight of the linear transformation.
Afterward, we combine $\mathbf{F}^\text{traj}$ and $\mathbf{F}^\text{dete}$ into a feature matrix $\mathbf{F} \in \mathbb{R}^{(S\times U) \times 2D}$, which models the spatial distribution pattern and velocity-time correlation, \textit{etc}.
A fully connected layer and a sigmoid function are then applied to yield the correlation score.
Here, we obtain the correlation matrix ${\mathbf{C}^{\text{corr}} \in \mathbb{R}^{S \times U}}$ reflecting the association probabilities between lost tracklets and unmatched detections.
Finally, we use the greedy algorithm to pick the matched pairs with high correlation socre and initialize the remaining unmatched detections as new tracklets.

\noindent {\bf Error Compensation.}
After re-identifying the lost tracklet with its matched detection, we need to fill the trajectory during the long-time occlusion.
Unlike other interpolation methods, we correct the predicted trajectory instead of generating a new one.
We use the error between the matched detection $\mathbf{d}_{}^{t}$ and the prediction $\mathbf{p}_{}^{t}$ of the lost tracklet to infer the errors during occlusion and refine the prediction:
\begin{equation}
\begin{aligned}
\mathbf{d}^{t_p} = \mathbf{p}^{t_p} + \left(\mathbf{d}_{}^{t}-\mathbf{p}_{}^{t}\right) \frac{t_p-t_1}{t_2-t_1},\ \ \   t_1<t_p<t_2, \\
\end{aligned}
\end{equation}
where the tracklet becomes lost after frame $t_1$ and is refound at frame $t_2$.  When the tracklet is occluded, it still exists and interacts with other tracklets, which can provide some information to support the prediction of the lost one. 
Related experiments are in the supplementary material.


\subsection{Training}
The output of the Interaction Module is a set of offsets $\mathbf{P}^{\text{offs}}$. We convert it to location coordinates $\mathbf{P}^{\text{coor}}$ based on location in the previous frame.
Each coordinate in $\mathbf{P}^{\text{coor}}$ is composed of four values $(x, y, w, h)$, which are not independent. 
These four variables affect each other, so they should not be supervised individually but combined to drive the training of the Interaction Module.
Inspired by the training process of the detector~\cite{ge2021yolox}, we employ the IoU loss~\cite{iou} to supervise the Interaction Module as follows:
\begin{equation}
\begin{aligned}
\mathcal{L}^{\text{INTR}} =1-\rm{IoU}\left(\mathbf{P}^{\text{coor}}, \mathbf{P}^\text{gt}\right)^2,
\end{aligned}
\end{equation}
where $\rm{IoU}$ denotes the Intersection over Union.
We take three consecutive frames as a training sample. The offset between the first two frames is taken as input, and the last frame is used as supervision $\mathbf{P}^\text{gt}$.

\begin{table*}[t]
\centering  
\resizebox{0.95\linewidth}{!}{
\setlength{\tabcolsep}{0.9em}%
\begin{tabular}{l|c|ccccccccc}
\toprule[2pt]
\textbf{Tracker} & \textbf{Venue} & \textbf{IDF1 $\uparrow$} & \textbf{MOTA  $\uparrow$} & \textbf{HOTA $\uparrow$} & \textbf{AssA $\uparrow$} & \textbf{DetA  $\uparrow$} & \textbf{FP $ \downarrow$} & \textbf{FN  $\downarrow$} & \textbf{IDs  $\downarrow$} & \textbf{Frag  $\downarrow$} \\ \hline
ReMOT~\cite{remot}             & IVC'21   & 72.0            & 77.0             & 59.7            & 57.1            & 62.8             & 33204            & 93612            & 2853              & 5304               \\
QuasiDense~\cite{pang2021quasi}        & CVPR'21  & 66.3            & 68.7             & 53.9            & 52.7            & 55.6             & 26589            & 146643           & 3378              & 8091               \\
SOTMOT~\cite{sotmot}            & CVPR'21  & 71.9            & 71.0             & -               & -               & -                & 39537            & 118983           & 5184              & -                  \\
SiamMOT~\cite{siammot}           & CVPR'21  & 72.3            & 76.3             & -               & -               & -                & -                & -                & -                 & -                  \\
CorrTracker~\cite{CorrTracker}       & CVPR'21  & 73.6            & 76.5             & 60.7            & 58.5            & 62.9             & 29808            & 99510            & 3369              & 6063               \\
PermaTrackPr~\cite{permatrackpr}      & ICCV'21  & 68.9            & 73.8             & 55.5            & 53.1            & 58.5             & 28998            & 115104           & 3699              & 6132               \\
FairMOT~\cite{fairmot}           & IJCV'21  & 72.3            & 73.7             & 59.3            & 58.0            & 60.9             & 27507            & 117477           & 3303              & 8073               \\
CSTrack~\cite{cstrack}           & TIP'22   & 72.6            & 74.9             & 59.3            & 57.9            & 61.1             & 23847            & 114303           & 3567              & 7668               \\
RelationTrack~\cite{relationtrack}     & TMM'22   & 74.7            & 73.8             & 61.0            & 61.5            & 60.6             & 27999            & 118623           & 1374              & 2166               \\
TrackFormer~\cite{trackformer}       & CVPR'22  & 68.0            & 74.1             & -               & -               & -                & 34602            & 108777           & 2829              & -                  \\
MeMOT~\cite{memot}             & CVPR'22  & 69.0            & 72.5             & 56.9            & 55.2            & -                & 37221            & 115248           & 2724              & -                  \\
MTrack~\cite{Towards-Discriminative}            & CVPR'22  & 73.5            & 72.1             & -               & -               & -                & 53361            & 101844           & 2028              & -                  \\
MOTR~\cite{motr}         & ECCV'22  & 68.6            & 73.4             & 57.8           & 55.7            & 60.3             & -           & -            & 2439             & -            \\
ByteTrack~\cite{bytetrack}         & ECCV'22  & 77.3            & 80.3             & 63.1            & 62.0            & 64.5             & 25491            & 83721            & 2196              & 2277               \\
P3AFormer(+W\&B)~\cite{p3aformer}  & ECCV'22  & 78.1            & \textbf{81.2}             & -               & -               & -                & \textbf{17281}            & 86861            & 1893              & -                  \\
\rowcolor[HTML]{DAE8FC} 
MotionTrack(ours) &     -      & \textbf{80.1}            & 81.1             & \textbf{65.1}            & \textbf{65.1}            & \textbf{65.4}             & 23802            & \textbf{81660}            & \textbf{1140}              & \textbf{1605}               \\ \bottomrule[2pt]
\end{tabular}}
\caption{Comparison with the state-of-the-art methods under the “private detector” protocol on the MOT17 test set. $\uparrow$ means higher is better, $\downarrow$ means lower is better. The best results for each metric are bolded. }  
\label{table1}  
\end{table*}

For the training of Refind Module, we extract the data samples and labels from the tracking dataset.
We first extract all the trajectories in a complete video, and then randomly couple them in pairs, each being a training set.
For each training set, we sample the tracklet and detection, and label positive or negative by whether they are from the same trajectory.
Then, we supervise the correlation calculation in Refind Module with a binary cross-entropy loss function:
\begin{equation}
\begin{aligned}
\mathcal{L}^{\text{CORR}}=\frac{1}{n} \sum_i^{n}-[y_i \log (c_i)+(1-y_i)\log(1-c_i)],
\end{aligned}
\end{equation}
where $c_i$ denotes the predicted correlation score, and $y_i$ indicates the ground truth correlation label, in which 1 and 0 represent the positive and negative correlation, respectively.

\section{Experiments}
\label{sec:Experiments}
\subsection{Setting}
\noindent {\bf Datasets.}
We evaluate our MotionTrack under the ``private detection"  protocol on the MOT17~\cite{mot16} and MOT20~\cite{mot20} datasets. For fair comparisons, we directly apply the publicly available detector of YOLOX~\cite{ge2021yolox}, trained by ByteTrack~\cite{bytetrack} on MOT17 and MOT20. 
For the training of the Interaction Module and Refind Module, 
we use only half the training set of MOT17 and MOT20.

\noindent {\bf Metrics.}
We employ CLEAR metrics~\cite{clear-metric} (MOTA, FP, FN, IDs, \textit{etc.}), IDF1~\cite{idf1}, and HOTA~\cite{hota} to evaluate different aspects of tracking performance. In particular,
IDF1 focuses more on association performance,
MOTA is computed based on FP, FN, and IDs, which mainly rely on the detection performance because the number of FPs and FNs is larger than IDs. HOTA is a unified metric that balances the effectiveness of detection and association.

\noindent {\bf Implementation Details.}
We implemented our MotionTrack in PyTorch~\cite{paszke2019pytorch}, and performed all experiments on one NVIDIA GeForce RTX 3090 Ti GPU. For fair comparisons, we directly apply the publicly available detector of YOLOX~\cite{ge2021yolox}, trained by~\cite{bytetrack} for MOT17, MOT20.
For the Interaction Module, the threshold used by the signal function is set to 0.6. For Refind Module, pairs with correlation scores less than 0.9 were rejected.
For the tracking process, we make full use of all detections with double matching, following~\cite{bytetrack}. The default high and low thresholds are 0.6 and 0.1, respectively. Unless otherwise specified, the new tracklet initialization score is 0.7. 
In the IoU-based association, we reject the matching if the IoU score is smaller than 0.2, and we use global motion compensation
for steadier tracking. For the lost tracklets, we keep 60 frames for MOT17 and 120 frames for MOT20, respectively.

\begin{table*}[t]
\centering  
\resizebox{0.95\linewidth}{!}{
\setlength{\tabcolsep}{0.9em}%
\begin{tabular}{l|c|ccccccccc}
\toprule[2pt]
\textbf{Tracker} & \textbf{Venue} & \textbf{IDF1 $\uparrow$} & \textbf{MOTA  $\uparrow$} & \textbf{HOTA $\uparrow$} & \textbf{AssA $\uparrow$} & \textbf{DetA  $\uparrow$} & \textbf{FP $ \downarrow$} & \textbf{FN  $\downarrow$} & \textbf{IDs  $\downarrow$} & \textbf{Frag  $\downarrow$} \\ \hline
FairMOT~\cite{fairmot}           & IJCV'21  & 67.3            & 61.8             & 54.6            & 54.7            & 54.7             & 103440           & 88901            & 5243              & 7874               \\
CorrTracker~\cite{CorrTracker}       & CVPR'21  & 69.1            & 65.2             & -               & -               & -                & 79429            & 95855            & 5183              & -                  \\
SiamMOT~\cite{siammot}           & CVPR'21  & 69.1            & 67.1             & -               & -               & -                & -                & -                & -                 & -                  \\
SOTMOT~\cite{sotmot}            & CVPR'21  & 71.4            & 68.6             & 57.4            & 57.3            & 57.7             & 57064            & 101154           & 4209              & 7568               \\

CSTrack~\cite{cstrack}           & TIP'22   & 68.6            & 66.6             & 54.0            & 50.0            & 54.2             & \textbf{25404}            & 144358           & 3196              & 7632               \\
RelationTrack~\cite{relationtrack}     & TMM'22   & 70.5            & 67.2             & 56.5            & 56.4            & 56.8             & 61134            & 104597           & 4243              & 8236               \\
MeMOT~\cite{memot}             & CVPR'22  & 66.1            & 63.7             & 54.1            & 55.0            & -                & 47882            & 137982           & 1938              & -                  \\
MTrack~\cite{Towards-Discriminative}            & CVPR'22  & 69.2            & 63.5             & -               & -               & -                & 96123            & 86964            & 6031              & -                  \\
ByteTrack~\cite{bytetrack}         & ECCV'22  & 75.2            & 77.8             & 61.3            & 59.6            & 63.4             & 26249            & 87594            & 1223              & 1460               \\
P3AFormer(+W\&B)~\cite{p3aformer}  & ECCV'22        & 76.4                     & \textbf{78.1}             & -                        & -                        & -                & 25413            & 86510            & 1332              & -                  \\
\rowcolor[HTML]{DAE8FC} 
MotionTrack(ours) &      -           & \textbf{76.5}            & 78.0                      & \textbf{62.8}            & \textbf{61.8}            & \textbf{64.0}    & 28629            & \textbf{84152}   & \textbf{1165}     & \textbf{1321}      \\ 
\bottomrule[2pt]
\end{tabular}
}
\caption{Comparison with the state-of-the-art methods under the “private detector” protocol on the MOT20 test set. $\uparrow$ means higher is better, $\downarrow$ means lower is better. The best results for each metric are bolded.}
\label{table2}  
\end{table*}

\subsection{Comparison with the State-of-the-Art Methods}

\noindent {\bf MOT17.}
As shown in~\cref{table1}, our MotionTrack outperforms the state-of-the-art methods in most key metrics, \emph{i.e.}, ranks first for metrics IDF1, HOTA, AssA, DetA, IDs, Frag and ranks second for MOTA. Our approach focuses on solving dense crowds and extreme occlusion to enable the tracker with a more robust ability for identity preservation over a short to long range. Consequently, it produces more accurate associations and vastly outperforms the second-performance tracker in metrics reflecting the association ability~(\emph{i.e.}, +2.0 IDF1 and +3.1 AssA). It should be pointed out that P3AFormer is based on segmentation, while our MotionTrack is based on detection. Even in this situation, we are only 0.1 lower than P3AFormer in MOTA, but IDF1 surpasses it by 2.0. 
We are the highest on HOTA, demonstrating the robustness and comprehensive tracking capability of our MotionTrack. For the IDs metric, we are 40\% less than P3AFormer, which shows the strong ability of our association component.

\begin{table}[t]
\centering
\renewcommand{\arraystretch}{1.1}
\setlength{\tabcolsep}{8pt}
\Huge
\resizebox{0.9\linewidth}{!}{
\begin{tabular}{l|cccccc}
\toprule[3.5pt]
\textbf{Setting}   & \textbf{IDF1  $\uparrow$}                           & \textbf{MOTA $\uparrow$}                            & \textbf{HOTA  $\uparrow$}                           & \textbf{AssA $ \uparrow$}                           & \textbf{DetA  $\uparrow$}  & \textbf{IDs  $\downarrow$}                          \\ \hline
Baseline     & 82.6            & 80.4            & 70.2            & 72.4            & 68.7            & 402            \\
Baseline+I   & 83.0            & 80.5            & 70.5            & 72.9            & 68.8            & 390            \\
Baseline+I+R & 83.7            & 80.7            & 70.8            & 73.5            & 68.9            & 378            \\ 
\bottomrule[3.5pt]
\end{tabular}}
\caption{Ablation studies on Interaction Module~(I) and Refind Module~(R) on the MOT17 validation set.}  
\label{table3}  
\end{table}

\begin{table}[t]
\centering
\renewcommand{\arraystretch}{1.1}
\setlength{\tabcolsep}{2.5pt}
\Huge
\resizebox{0.9\linewidth}{!}{
\setlength{\tabcolsep}{0.5em}%
\begin{tabular}{l|cccccc}
\toprule[3.2pt]
\textbf{Setting}    & \textbf{\#} & \textbf{IDF1  $\uparrow$}                           & \textbf{MOTA $\uparrow$}                            & \textbf{HOTA  $\uparrow$}                           & \textbf{AssA $ \uparrow$}                           & \textbf{DetA  $\uparrow$}                                             \\ \hline
           & 30          & 80.9          & 79.8          & 69.0          & 70.6          & 68.1 \\
IoU-based  & 120         & 80.1          & 77.6          & 68.4          & 70.5          & 66.9 \\ \cline{2-7} 
           &  $\Delta$           & \textcolor{red}{\bf -0.8} & \textcolor{red}{\bf -2.2} & \textcolor{red}{\bf -0.6} & \textcolor{red}{\bf -0.1} & \textcolor{red}{\bf -1.2}\\\hline
           & 30          & 77.2          & 77.0          & 66.4          & 67.1          & 66.3 \\
ReID-based & 120         & 70.4          & 67.5          & 60.6          & 60.3          & 61.6 \\ \cline{2-7} 
           &   $\Delta$            & \textcolor{red}{\bf -6.8} & \textcolor{red}{\bf -9.5} & \textcolor{red}{\bf -5.8} & \textcolor{red}{\bf -6.8} & \textcolor{red}{\bf -4.7}\\\hline
           & 30          & 82.6          & 80.4          & 70.2          & 72.4          & 68.7 \\
Ours       & 120         & 83.3          & 80.7          & 70.7          & 73.2          & 68.8 \\ \cline{2-7} 
           &   $\Delta$            & \textcolor[RGB]{0,204,0}{\bf \bf +0.7} & \textcolor[RGB]{0,204,0}{\bf +0.3} & \textcolor[RGB]{0,204,0}{\bf +0.5} & \textcolor[RGB]{0,204,0}{\bf \bf +0.8} & \textcolor[RGB]{0,204,0}{\bf +0.1}\\
\bottomrule[3.2pt]
\end{tabular}}
\caption{Comparison with other methods for handling occlusions on the MOT17 validation set. We raise the upper limit of occlusion time from 30 to 120 frames to reflect the ability to deal with long-term occlusion. Increases and decreases in metrics are marked in green and red, respectively.}  
\label{table4}  
\end{table}

\begin{table}[t]
\centering
\renewcommand{\arraystretch}{1.1}
\setlength{\tabcolsep}{2.5pt}
\Huge
\resizebox{0.9\linewidth}{!}{
\setlength{\tabcolsep}{1.1em}%
\begin{tabular}{l|cccccc}
\toprule[3pt]
\textbf{Setting}  & \textbf{$\geqslant$ 20} & \textbf{$\geqslant$ 40} & \textbf{$\geqslant$ 60} & \textbf{$\geqslant$ 80}  & \textbf{$\geqslant$ 100}  \\ \hline
Baseline & 77.2        & 73.6        & 75.2        & 73.3        & 71.5         \\
Ours     & 78.3        & 75.1        & 76.8        & 74.1        & 72.4         \\ \hline
Improvement & \textcolor[RGB]{0,204,0}{\bf +1.1} & \textcolor[RGB]{0,204,0}{\bf +1.5} & \textcolor[RGB]{0,204,0}{\bf +1.6} & \textcolor[RGB]{0,204,0}{\bf +0.8} & \textcolor[RGB]{0,204,0}{\bf +0.9} \\ 
\bottomrule[3pt]
\end{tabular}}
\caption{Evaluation of MOTA for crowd and occlusion cases on the MOT17 validation set. We set visibility~$<$~0.25 and define the minimum time constants for occlusion or crowds as 20 to 100, respectively. Increases and decreases in metrics are marked in green and red, respectively.}  
\label{table5}  
\end{table}

\noindent {\bf MOT20.} 
As shown in~\cref{table2}, our MotionTrack still achieves state-of-the-art results on the MOT20 dataset~$^{1}$~\footnote{
$^{1}$ We ranked second among all methods on the official MOT Challenge evaluation server and first among all online methods.}. 
Even though ByteTrack uses duplicate detections as our method, our MotionTrack has made significant progress in several core metrics~(\emph{i.e.}, +1.3 IDF1, +0.2 MOTA, and +1.5 HOTA). Besides, we still achieve the highest HOTA in this more challenging scenario.
Moreover, our MotionTrack achieves the least IDs, which is 13\% less than the second P3AFormer. 
The underlying reason is that we infer the occluded trajectory in the Refind Module, which is able to keep a consistent identity for each trajectory.

\subsection{Ablation Study}
\noindent {\bf Effect of Each Component.}
We conduct ablative experiments to verify the contribution of each component of our MotionTrack. For reliable verification, all other settings of baseline are the same, except it utilizes the Kalman Filter~(KF) for motion prediction and re-identifies the lost tracklet using the IoU score between the tracklet’s iterative prediction and detection. Then, our proposed Interaction Module and Refind Module replace the above two components, respectively. As shown in~\cref{table3}, the Interaction Module can improve IDF1, HOTA, AssA, and IDs, which indicates the effectiveness of the introduction of interaction. Our Refind Module can improve all the metrics, indicating that the learned history trajectory-based motions are effective for the long-range data association problem.

\begin{figure}[t]
  \centering
  \includegraphics[width=0.98\linewidth]{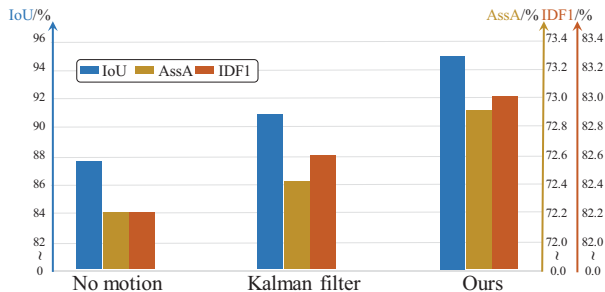}
   \caption{{\bf Results of Motion Models}. The three metrics reflect the motion prediction accuracy (IoU) and the ability of association (IDF1 and AssA), respectively.}
   \label{fig:histogram}
\end{figure}

\noindent {\bf Analysis of Interation Module.}
The IoU-based trackers predict the location of tracklets in the next frame, and then compute the IoU between the detection boxes and the predicted boxes to conduct data association. These methods heavily rely on accurate predictions to maintain a high-quality data association. As shown in Figure~\ref{fig:histogram}, we compare the prediction accuracy, in which we take the average IoU of all targets to count the total IoU score. The results show that the introduction of interaction yields more accurate predictions than the traditional Kalman Filter. Meanwhile, steady improvements in IDF1 and AssA indicate that Interaction Module improves prediction accuracy and leads to stronger association capability.

\noindent {\bf Advantage of Refind Module.} In practice, the IoU-based and Re-ID-based approaches are often taken to deal with the occlusion problem, while our method learns the discriminative motion patterns between the history trajectories and current detection to address the extreme occlusion problem. As shown in~\cref{table4}, we compare the classical approach~\cite{bytetrack,deepsort} with our method in terms of its ability to handle long-term occlusion. The results show that the first two methods cause dramatic performance decreases due to a large number of incorrect 
associations, while our approach method achieves a consistent improvement in dealing the long-term occlusion.

\noindent {\bf Extra Evaluation of Crowd and Occlusion.} Extreme occlusion and dense crowds are not always present in the tracking scene. Therefore evaluating the entire dataset cannot reflect our ability to directly solve the two challenging problems. To address this issue, we propose to modify MOTA to crowdMOTA, which separates people in the dataset who are likely to be occluded or crowded, and evaluate the tracking performance of these people. In particular, we select people by visibility label in the dataset, \textit{i.e.}, considering people being crowded or occluded when their visibility is below 0.25.
We set the minimum number of frames from 20 to 100 for continuous occlusion or crowding to validate the solution for samples with different difficulty levels.
As shown inc~\cref{table5}, the improvement in samples of different difficulties illustrates the effectiveness of our method in solving dense crowds and extreme occlusion.

\subsection{Visualization}
\begin{figure}[t]
  \centering
  \includegraphics[width=1\linewidth]{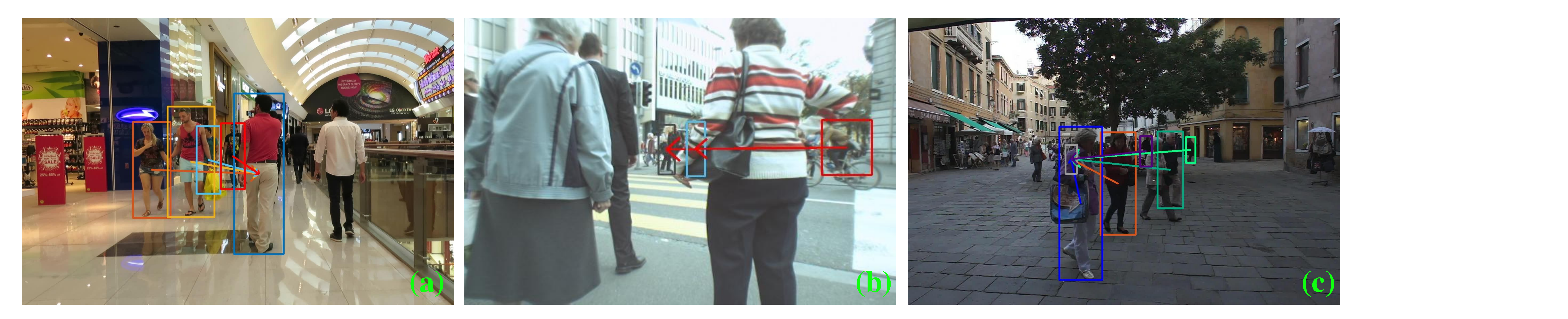}
   \caption{{\bf  Visualization of Directed Interaction}. The boxes in different colors represent the bounding boxes of different targets, the end of arrow represents the affecting target, and the head of arrow represents affected target.}
   \label{fig:visual1}
\end{figure}

\noindent {\bf Visualization of Directed Interaction.}
The directed interaction is visualized in Figure~\ref{fig:visual1}, from which we find that our method possesses the ability to capture effective interaction in different scenarios. In particular,~(a) shows that the stride forward movement pattern of the pedestrian is affected by the pedestrians coming towards him. In~(b), a cyclist walking toward the left will affect two oncoming pedestrians. As shown in~(c), even if the tracklet is in the LOST state, he is still influenced by the pedestrians around him. Although he cannot be seen, he is still actually in the crowd. As a result, considering the interaction of nearby people helps describe the movement of the target during occlusion.

\begin{figure}[t]
  \centering
  \includegraphics[width=1\linewidth]{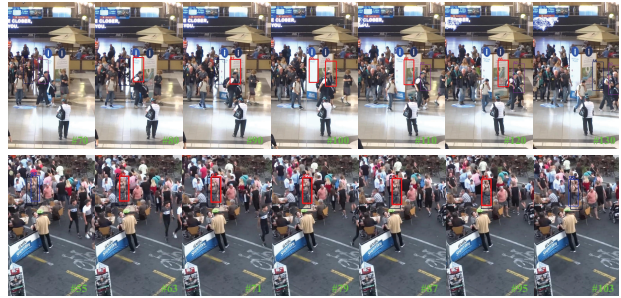}
   \caption{{\bf Visualization of Refinding Targets}. The red boxes represent the locations during occlusion, which are obtained based on both the iterative prediction of the Interaction Module and the error compensation of the Refind Module.
}
   \label{fig:visual2}
\end{figure}

\noindent {\bf Visualization of Refinding Targets.} 
As shown in Figure~\ref{fig:visual2}, two cases of refinding targets are given in the red box. When the target is occluded, our Interaction Module still iteratively infers its location. Therefore, our Refind Module can accurately re-identify the lost targets through correlation calculation and refine the predicted trajectory by error compensation.

\section{Conclusion}
\label{sec:Conclusion}
We propose MotionTrack for online MOT, which introduces an Interaction Module and Refind Module to address the short-term and long-term association problems in MOT. Our results on the MOT benchmark datasets have shown the benefits of our method. The application of interactions can lead to more accurate location prediction, yielding more robust data association. Even though the targets have been occluded for a long period, they can still be refound by learning discriminative motion patterns between the history trajectory and current detection. Notably, without using any complex components, such as person Re-ID, our tracker achieves state-of-the-art performance.

\noindent \textbf{Limitation and Future Work.} One major limitation of our method is that we only considered the motion patterns and relationships between pedestrians while ignoring the drivable information in interaction, which may weaken its performance in motion prediction. In the future, we plan to explore this prior information in our MotionTrack to further improve the tracking performance. Besides, two modules we proposed can be further combined to support each other for achieving better results.

\section*{Acknowledgement}
This work was supported in part by National Key R\&D Program of China under Grant 2021YFB1714700, NSFC under Grants 62088102 and 62106192, Natural Science Foundation of Shaanxi Province under Grants 2022JC-41, China Postdoctoral Science Foundation under Grant 2022T150518, and Fundamental Research Funds for the Central Universities under Grants XTR042021005 and XTR072022001.

{\small
\bibliographystyle{ieee_fullname}
\bibliography{egbib}

\begin{thebibliography}{10}\itemsep=-1pt

\bibitem{botsort}
Nir Aharo, Roy Orfaig, and Ben-Zion Bobrovsky.
\newblock {BoT-SORT}: Robust associations multi-pedestrian tracking.
\newblock {\em arXiv preprint arXiv:2206.14651}, 2022.

\bibitem{tracking-by-detection}
Mykhaylo Andriluka, Stefan Roth, and Bernt Schiele.
\newblock People-tracking-by-detection and people-detection-by-tracking.
\newblock In {\em CVPR}, pages 1--8, 2008.

\bibitem{tracktor}
Philipp Bergmann, Tim Meinhardt, and Laura Leal-Taixe.
\newblock Tracking without bells and whistles.
\newblock In {\em ICCV}, pages 941--951, 2019.

\bibitem{clear-metric}
Keni Bernardin and Rainer Stiefelhagen.
\newblock Evaluating multiple object tracking performance: the clear mot
  metrics.
\newblock {\em JIVP}, 2008:1--10, 2008.

\bibitem{sort}
Alex Bewley, Zongyuan Ge, Lionel Ott, Fabio Ramos, and Ben Upcroft.
\newblock Simple online and realtime tracking.
\newblock In {\em ICIP}, pages 3464--3468, 2016.

\bibitem{braso2020learning}
Guillem Bras{\'o} and Laura Leal-Taix{\'e}.
\newblock Learning a neural solver for multiple object tracking.
\newblock In {\em CVPR}, pages 6247--6257, 2020.

\bibitem{memot}
Jiarui Cai, Mingze Xu, Wei Li, Yuanjun Xiong, Wei Xia, Zhuowen Tu, and Stefano
  Soatto.
\newblock {MeMOT}: Multi-object tracking with memory.
\newblock In {\em CVPR}, pages 8090--8100, 2022.

\bibitem{driving}
Chenyi Chen, Ari Seff, Alain Kornhauser, and Jianxiong Xiao.
\newblock {DeepDriving}: Learning affordance for direct perception in
  autonomous driving.
\newblock In {\em ICCV}, pages 2722--2730, 2015.

\bibitem{mot20}
Patrick Dendorfer, Hamid Rezatofighi, Anton Milan, Javen Shi, Daniel Cremers,
  Ian Reid, Stefan Roth, Konrad Schindler, and Laura Leal-Taix{\'e}.
\newblock {MOT20}: A benchmark for multi object tracking in crowded scenes.
\newblock {\em arXiv preprint arXiv:2003.09003}, 2020.

\bibitem{dendorfer2022quo}
Patrick Dendorfer, Vladimir Yugay, Aljo{\v{s}}a O{\v{s}}ep, and Laura
  Leal-Taix{\'e}.
\newblock Quo vadis: Is trajectory forecasting the key towards long-term
  multi-object tracking?
\newblock 2022.

\bibitem{ding2019acnet}
Xiaohan Ding, Yuchen Guo, Guiguang Ding, and Jungong Han.
\newblock {ACNet}: Strengthening the kernel skeletons for powerful cnn via
  asymmetric convolution blocks.
\newblock In {\em ICCV}, pages 1911--1920, 2019.

\bibitem{strongsort}
Yunhao Du, Yang Song, Bo Yang, and Yanyun Zhao.
\newblock {StrongSORT}: Make deepsort great again.
\newblock {\em arXiv preprint arXiv:2202.13514}, 2022.

\bibitem{giaotracker}
Yunhao Du, Junfeng Wan, Yanyun Zhao, Binyu Zhang, Zhihang Tong, and Junhao
  Dong.
\newblock {GIAOTracker}: A comprehensive framework for mcmot with global
  information and optimizing strategies in visdrone 2021.
\newblock In {\em ICCV}, pages 2809--2819, 2021.

\bibitem{ge2021yolox}
Zheng Ge, Songtao Liu, Feng Wang, Zeming Li, and Jian Sun.
\newblock {YOLOX}: Exceeding yolo series in 2021.
\newblock {\em arXiv preprint arXiv:2107.08430}, 2021.

\bibitem{Online-multiple}
Song Guo, Jingya Wang, Xinchao Wang, and Dacheng Tao.
\newblock Online multiple object tracking with cross-task synergy.
\newblock In {\em CVPR}, pages 8136--8145, 2021.

\bibitem{socialgan}
Agrim Gupta, Justin Johnson, Li Fei-Fei, Silvio Savarese, and Alexandre Alahi.
\newblock Social gan: Socially acceptable trajectories with generative
  adversarial networks.
\newblock In {\em CVPR}, pages 2255--2264, 2018.

\bibitem{Motion-aware}
Shoudong Han, Piao Huang, Hongwei Wang, En Yu, Donghaisheng Liu, and Xiaofeng
  Pan.
\newblock {MAT}: Motion-aware multi-object tracking.
\newblock {\em Neurocomputing}, 476:75--86, 2022.

\bibitem{he2021learnable}
Jiawei He, Zehao Huang, Naiyan Wang, and Zhaoxiang Zhang.
\newblock Learnable graph matching: Incorporating graph partitioning with deep
  feature learning for multiple object tracking.
\newblock In {\em CVPR}, pages 5299--5309, 2021.

\bibitem{kalman1960new}
Rudolph~Emil Kalman.
\newblock A new approach to linear filtering and prediction problems.
\newblock 1960.

\bibitem{Detecting-invisible-people}
Tarasha Khurana, Achal Dave, and Deva Ramanan.
\newblock Detecting invisible people.
\newblock In {\em ICCV}, pages 3174--3184, 2021.

\bibitem{kim2015multiple}
Chanho Kim, Fuxin Li, Arridhana Ciptadi, and James~M Rehg.
\newblock Multiple hypothesis tracking revisited.
\newblock In {\em CVPR}, pages 4696--4704, 2015.

\bibitem{kipf2016semi}
Thomas~N Kipf and Max Welling.
\newblock Semi-supervised classification with graph convolutional networks.
\newblock {\em arXiv preprint arXiv:1609.02907}, 2016.

\bibitem{siammot}
Chao Liang, Zhipeng Zhang, Xue Zhou, Bing Li, and Weiming Hu.
\newblock One more check: making “fake background” be tracked again.
\newblock In {\em AAAI}, pages 1546--1554, 2022.

\bibitem{cstrack}
Chao Liang, Zhipeng Zhang, Xue Zhou, Bing Li, Shuyuan Zhu, and Weiming Hu.
\newblock Rethinking the competition between detection and reid in multiobject
  tracking.
\newblock {\em IEEE T-IP}, 31:3182--3196, 2022.

\bibitem{hota}
Jonathon Luiten, Aljosa Osep, Patrick Dendorfer, Philip Torr, Andreas Geiger,
  Laura Leal-Taix{\'e}, and Bastian Leibe.
\newblock {HOTA}: A higher order metric for evaluating multi-object tracking.
\newblock {\em IJCV}, 129(2):548--578, 2021.

\bibitem{trackformer}
Tim Meinhardt, Alexander Kirillov, Laura Leal-Taixe, and Christoph
  Feichtenhofer.
\newblock {TrackFormer}: Multi-object tracking with transformers.
\newblock In {\em CVPR}, pages 8844--8854, 2022.

\bibitem{mot16}
Anton Milan, Laura Leal-Taix{\'e}, Ian Reid, Stefan Roth, and Konrad Schindler.
\newblock {MOT16}: A benchmark for multi-object tracking.
\newblock {\em arXiv preprint arXiv:1603.00831}, 2016.

\bibitem{munkres1957algorithms}
James Munkres.
\newblock Algorithms for the assignment and transportation problems.
\newblock {\em Journal of the Society for Industrial and Applied Mathematics},
  5(1):32--38, 1957.

\bibitem{surveil}
Sangmin Oh, Anthony Hoogs, Amitha Perera, Naresh Cuntoor, Chia-Chih Chen,
  Jong~Taek Lee, Saurajit Mukherjee, JK Aggarwal, Hyungtae Lee, Larry Davis,
  et~al.
\newblock A large-scale benchmark dataset for event recognition in surveillance
  video.
\newblock In {\em CVPR}, pages 3153--3160, 2011.

\bibitem{pang2021quasi}
Jiangmiao Pang, Linlu Qiu, Xia Li, Haofeng Chen, Qi Li, Trevor Darrell, and
  Fisher Yu.
\newblock Quasi-dense similarity learning for multiple object tracking.
\newblock In {\em CVPR}, pages 164--173, 2021.

\bibitem{paszke2019pytorch}
Adam Paszke, Sam Gross, Francisco Massa, Adam Lerer, James Bradbury, Gregory
  Chanan, Trevor Killeen, Zeming Lin, Natalia Gimelshein, Luca Antiga, et~al.
\newblock {PyTorch}: An imperative style, high-performance deep learning
  library.
\newblock In {\em NeurIPS}, pages 8026--8037, 2019.

\bibitem{faster-rcnn}
Shaoqing Ren, Kaiming He, Ross Girshick, and Jian Sun.
\newblock {Faster R-CNN}: Towards real-time object detection with region
  proposal networks.
\newblock In {\em NIPS}, pages 91--99, 2015.

\bibitem{idf1}
Ergys Ristani, Francesco Solera, Roger Zou, Rita Cucchiara, and Carlo Tomasi.
\newblock Performance measures and a data set for multi-target, multi-camera
  tracking.
\newblock In {\em ECCV}, pages 17--35, 2016.

\bibitem{probabilistic}
Fatemeh Saleh, Sadegh Aliakbarian, Hamid Rezatofighi, Mathieu Salzmann, and
  Stephen Gould.
\newblock Probabilistic tracklet scoring and inpainting for multiple object
  tracking.
\newblock In {\em CVPR}, pages 14329--14339, 2021.

\bibitem{shi2021sgcn}
Liushuai Shi, Le Wang, Chengjiang Long, Sanping Zhou, Mo Zhou, Zhenxing Niu,
  and Gang Hua.
\newblock Sgcn: Sparse graph convolution network for pedestrian trajectory
  prediction.
\newblock In {\em CVPR}, pages 8994--9003, 2021.

\bibitem{modelling-ambiguous}
Daniel Stadler and J{\"u}rgen Beyerer.
\newblock Modelling ambiguous assignments for multi-person tracking in crowds.
\newblock In {\em WACV}, pages 133--142, 2022.

\bibitem{permatrackpr}
Pavel Tokmakov, Jie Li, Wolfram Burgard, and Adrien Gaidon.
\newblock Learning to track with object permanence.
\newblock In {\em ICCV}, pages 10860--10869, 2021.

\bibitem{vaswani2017attention}
Ashish Vaswani, Noam Shazeer, Niki Parmar, Jakob Uszkoreit, Llion Jones,
  Aidan~N Gomez, {\L}ukasz Kaiser, and Illia Polosukhin.
\newblock Attention is all you need.
\newblock In {\em NIPS}, pages 6000--6010, 2017.

\bibitem{regression2}
Xingyu Wan, Jiakai Cao, Sanping Zhou, Jinjun Wang, and Nanning Zheng.
\newblock Tracking beyond detection: learning a global response map for
  end-to-end multi-object tracking.
\newblock {\em IEEE T-IP}, 30:8222--8235, 2021.

\bibitem{regression1}
Xingyu Wan, Sanping Zhou, Jinjun Wang, and Rongye Meng.
\newblock Multiple object tracking by trajectory map regression with temporal
  priors embedding.
\newblock In {\em ACM MM}, pages 1377--1386, 2021.

\bibitem{wang2021multiple}
Qiang Wang, Yun Zheng, Pan Pan, and Yinghui Xu.
\newblock Multiple object tracking with correlation learning.
\newblock In {\em CVPR}, pages 3876--3886, 2021.

\bibitem{CorrTracker}
Qiang Wang, Yun Zheng, Pan Pan, and Yinghui Xu.
\newblock Multiple object tracking with correlation learning.
\newblock In {\em CVPR}, pages 3876--3886, 2021.

\bibitem{wang2020towards}
Zhongdao Wang, Liang Zheng, Yixuan Liu, Yali Li, and Shengjin Wang.
\newblock Towards real-time multi-object tracking.
\newblock In {\em ECCV}, pages 107--122, 2020.

\bibitem{deepsort}
Nicolai Wojke, Alex Bewley, and Dietrich Paulus.
\newblock Simple online and realtime tracking with a deep association metric.
\newblock In {\em ICIP}, pages 3645--3649, 2017.

\bibitem{remot}
Fan Yang, Xin Chang, Sakriani Sakti, Yang Wu, and Satoshi Nakamura.
\newblock {ReMOT}: A model-agnostic refinement for multiple object tracking.
\newblock {\em Image Vis. Comput.}, 106:104091, 2021.

\bibitem{Towards-Discriminative}
En Yu, Zhuoling Li, and Shoudong Han.
\newblock Towards discriminative representation: Multi-view trajectory
  contrastive learning for online multi-object tracking.
\newblock In {\em CVPR}, pages 8834--8843, 2022.

\bibitem{yu2022towards}
En Yu, Zhuoling Li, and Shoudong Han.
\newblock Towards discriminative representation: Multi-view trajectory
  contrastive learning for online multi-object tracking.
\newblock In {\em CVPR}, pages 8834--8843, 2022.

\bibitem{relationtrack}
En Yu, Zhuoling Li, Shoudong Han, and Hongwei Wang.
\newblock {RelationTrack}: Relation-aware multiple object tracking with
  decoupled representation.
\newblock {\em IEEE T-MM}, 2022.

\bibitem{iou}
Jiahui Yu, Yuning Jiang, Zhangyang Wang, Zhimin Cao, and Thomas Huang.
\newblock {UnitBox}: An advanced object detection network.
\newblock In {\em ACM MM}, pages 516--520, 2016.

\bibitem{motr}
Fangao Zeng, Bin Dong, Tiancai Wang, Xiangyu Zhang, and Yichen Wei.
\newblock Motr: End-to-end multiple-object tracking with transformer.
\newblock {\em arXiv preprint arXiv:2105.03247}, 2021.

\bibitem{fft}
Jimuyang Zhang, Sanping Zhou, Xin Chang, Fangbin Wan, Jinjun Wang, Yang Wu, and
  Dong Huang.
\newblock Multiple object tracking by flowing and fusing.
\newblock {\em arXiv preprint arXiv:2001.11180}, 2020.

\bibitem{bytetrack}
Yifu Zhang, Peize Sun, Yi Jiang, Dongdong Yu, Fucheng Weng, Zehuan Yuan, Ping
  Luo, Wenyu Liu, and Xinggang Wang.
\newblock {ByteTrack}: Multi-object tracking by associating every detection
  box.
\newblock In {\em ECCV}, pages 1--21, 2022.

\bibitem{fairmot}
Yifu Zhang, Chunyu Wang, Xinggang Wang, Wenjun Zeng, and Wenyu Liu.
\newblock Fairmot: On the fairness of detection and re-identification in
  multiple object tracking.
\newblock {\em IJCV}, 129(11):3069--3087, 2021.

\bibitem{p3aformer}
Zelin Zhao, Ze Wu, Yueqing Zhuang, Boxun Li, and Jiaya Jia.
\newblock Tracking objects as pixel-wise distributions.
\newblock In {\em ECCV}, pages 76--94, 2022.

\bibitem{sotmot}
Linyu Zheng, Ming Tang, Yingying Chen, Guibo Zhu, Jinqiao Wang, and Hanqing Lu.
\newblock Improving multiple object tracking with single object tracking.
\newblock In {\em CVPR}, pages 2453--2462, 2021.

\bibitem{zhoureid2}
Sanping Zhou, Fei Wang, Zeyi Huang, and Jinjun Wang.
\newblock Discriminative feature learning with consistent attention
  regularization for person re-identification.
\newblock In {\em ICCV}, pages 8040--8049, 2019.

\bibitem{zhoureid3}
Sanping Zhou, Jinjun Wang, Deyu Meng, Yudong Liang, Yihong Gong, and Nanning
  Zheng.
\newblock Discriminative feature learning with foreground attention for person
  re-identification.
\newblock {\em IEEE T-IP}, 28:4671--4684, 2019.

\bibitem{zhoureid1}
Sanping Zhou, Jinjun Wang, Jiayun Wang, Yihong Gong, and Nanning Zheng.
\newblock Point to set similarity based deep feature learning for person
  re-identification.
\newblock In {\em CVPR}, pages 3741--3750, 2017.

\bibitem{centertrack}
Xingyi Zhou, Vladlen Koltun, and Philipp Kr{\"a}henb{\"u}hl.
\newblock Tracking objects as points.
\newblock In {\em ECCV}, pages 474--490, 2020.

\end{thebibliography}
}

\end{document}